\documentclass[letterpaper]{article}
\usepackage{aaai}
\usepackage{times}
\usepackage{helvet}
\usepackage{courier}
\usepackage{color}
\usepackage{amsmath,bm}
\usepackage{array}
\usepackage{multirow}
\usepackage{multicol} 
\usepackage{xfrac}
\usepackage{algorithmic}
\usepackage[inoutnumbered,linesnumbered,ruled,vlined]{algorithm2e}
\usepackage{amsfonts}
\usepackage[small]{caption}

\usepackage{graphicx}
\usepackage{mathrsfs}
\usepackage{epsfig}
\usepackage{epstopdf}
\usepackage{subfigure}
\usepackage{enumerate}
\usepackage{helvet}
\usepackage{url}
\usepackage{amsmath,amssymb,amsthm}
\usepackage{bbding}
\usepackage{pifont}
\usepackage{rotating}
\usepackage{mathtools}
\usepackage{enumitem}

\renewcommand{\l}{$\langle$}
\renewcommand{\r}{$\rangle$}

\newcommand{\Dx}[1]{$\mathcal{D}_{#1}$\,}
\newcommand{\T}{$\mathcal{T}$\,}

\newcommand{\A}{$\mathcal{A}$\,}

\newcommand{\elpp}{$\mathcal{EL}^{++}$\,}

\newcommand{\bOx}[1]{$\mathbb{O}_{#1}$\,}

\newcommand{\kr}[1]{{\color{black}#1}}

\newtheorem{defn}{Definition}

\newtheorem{prop}{Property}
\newtheorem{ex}{Example}
 
\newtheorem{lemma}{Lemma}

\begin{document}
%
\title{Knowledge-based Transfer Learning Explanation}
\author{
Jiaoyan Chen\\
Department of Computer Science\\
University of Oxford, UK
\And 
Freddy L\'ecu\'e\\
INRIA, France\\
Accenture Labs, Ireland
\And 
Jeff Z. Pan\\
Department of Computer Science\\
University of Aberdeen, UK
\AND
Ian Horrocks\\
Department of Computer Science\\
University of Oxford, UK
\And
Huajun Chen\\
College of Computer Science and Technology\\
Zhejiang University, China
}
\maketitle
\begin{abstract}
\begin{quote}
\kr{Machine learning explanation can significantly boost machine learning's application in decision making,
but the usability of current methods is limited in human-centric explanation,
especially for transfer learning,
an important machine learning branch that aims at utilizing knowledge from one learning domain (i.e., a pair of dataset and prediction task) to enhance prediction model training in another learning domain.
In this paper, we propose an ontology-based approach for human-centric explanation of transfer learning.
}
Three kinds of knowledge-based explanatory \kr{evidence}, with different granularities, including general factors, particular narrators and core contexts are first proposed
and then inferred with both local ontologies and external knowledge bases.
The evaluation with US flight data and DBpedia has presented their confidence and availability in explaining the transferability of feature representation in flight departure delay forecasting.
\end{quote}
\end{abstract}

\section{Introduction}
\kr{Prediction with machine learning (ML) has been increasingly applied in a variety of fields to assist humans in decision making.
ML explanation work such as interpreting the prediction model or justifying the prediction result can significantly increase 
decision makers' confidence on the prediction and boost its application \cite{biran2017explanation},
}
especially in making critical decisions like cancer diagnosis when people need to understand how and why the prediction is made.

Most ML explanation studies such as designing inherently interpretable models \cite{wu2017beyond} and approximating a ``black box" model with multiple ``white box" models \cite{ribeiro2016should} aim at users with ML expertise.
The explanations lack background and common sense knowledge, 
thus are too hard to be understood by \kr{non-ML-experts, 
those common users without ML expertise such as doctors.
There are only a limited number of human-centric ML explanation studies.
Most of them adopt some corpus (e.g., Wikipedia articles \cite{biran2017human}) or Link Data \cite{tiddi2014dedalo}
to generate text to describe model components (e.g., effective \kr{ML} features) or justify the prediction results (e.g., data clusters).}
They adopt background knowledge but are limited by expressivity, which in turn restricts the reasoning and inhibits rich explanations. 

On the other hand, transfer learning which  utilizes samples, features (i.e., representations of original data) or models of one learning domain (\kr{i.e., a pair of dataset and prediction task) to enhance prediction model training} in another learning domain \cite{pan2010survey}
has been widely applied,
especially in dealing with critical challenges like lacking training data.
\kr{Its explanation aims at justifying the good or bad performance of the prediction model trained by a specific transfer learning algorithm with a specific parameter setting.
Current work on transfer learning explanation such as analyzing the impact of a feature's specificity and generality on its transferability \cite{yosinski2014transferable} aim at ML experts and
represent the insights in a machine understandable way.
It's hard for common users to understand why transfer from one learning domain contributes to an accurate prediction model (i.e., positive transfer) while transfer from another learning domain contributes to an inaccurate prediction model (i.e., negative transfer). 
}

In this paper, we propose an ontology-based knowledge representation and reasoning framework for human-centric transfer learning explanation.
\kr{
It first models a learning domain in transfer learning, including the dataset and the prediction task, 
with expressive OWL (Web Ontology Language \cite{bechhofer2009owl}) ontologies, 
and then complements the learning domain with the prediction task-related common sense knowledge using an efficient individual matching and external knowledge importing algorithm.
The framework further uses a correlative reasoning algorithm to infer three kinds of explanatory evidence (i.e., general factors, particular narrators and core contexts) 
to explain a positive feature or a negative transfer from one learning domain to another.
Some technical challenges such as feature transferability measurement and core context (entailment subset) searching are overcome.
}

\kr{As far as we know this is the first work to study human-centric transfer learning explanation and ontology-based ML explanation.
It achieves confident and rich human understandable evidence for explaining both positive and negative transfers in predicting US flight delay,
where the feature learned by a Convolutional Neural Network (CNN) is transferred. 
For example, we find that transferring between flights carried a big airline company is an evidence to explain positive transfers, 
while transferring between flights departing from the airport of SFO is an evidence to explain negative transfers (cf. Example \ref{ex:ote} for more examples). 
}

\kr{The remainder of the paper is organized as follows.
The next section introduces the ML background with \kr{ontologies}, and defines the problem of transfer learning explanation.}
Then we present the ontology-based framework and report the evaluation.
In the final two sections, we review the related work and conclude the paper.
\section{Background and Problem Definition}
 \vspace{-0.08cm}
We use Description Logics (DL) based ontologies written in the W3C OWL 2 standard\footnote{https://www.w3.org/TR/owl2-overview/},
in particular the $\mathcal{EL}^{++}$ \cite{baader2005pushing} fragment of the OWL 2 EL profile. 
In this section, we
first introduce $\mathcal{EL}^{++}$ based ontology,
then revisit \kr{the notions of learning domain, supervised learning and transfer learning} with ontologies, 
and eventually define the problem of   transfer learning explanation. 
\subsection{The  $\mathcal{EL}^{++}$ Description Logic}
Given a signature $\Sigma=(\mathcal{N}_C, \mathcal{N}_R, \mathcal{N}_I)$, consisting of $3$ disjoint sets of atomic concepts $\mathcal{N}_C$, atomic roles $\mathcal{N}_R$, and  individuals $\mathcal{N}_I$,
the top concept $\top$, the bottom concept $\bot$, an atomic concept $A$, an individual $a$, an atomic role   $r$, $\mathcal{EL}^{++}$ concept expressions $C$ and $D$ can be composed with the following constructs:
\begin{equation}
\vspace{-0.03cm}
\top\;|\;\bot\;|\;A\;|\;C\sqcap D\;|\;\exists r.C\;|\;\{a\}\nonumber
\end{equation}

An $\mathcal{EL}^{++}$ ontology 
is composed of a TBox $\mathcal{T}$ and an ABox $\mathcal{A}$. 
The TBox \T is a set of concept and
role axioms. 
$\mathcal{EL}^{++}$ supports General Concept Inclusion axioms (GCIs e.g., $C \sqsubseteq D$), 
Role Inclusion axioms (RIs e.g., $r_1 \sqsubseteq r_2$, $r_1 \circ r_2\sqsubseteq s$), where $C$, $D$ are concept expressions, $r_1$, $r_2$, $s$ are atomic roles.
The ABox \A is a set of class assertion axioms e.g., $C(a)$, role assertion axioms e.g., $r(a, b)$, 
individual equality and inequality axioms e.g., $a = b$, $a \neq b$, where C is a concept expression, $r$ is an atomic roles and $a$, $b$ are individuals.
Entailment reasoning in \elpp is PTime-Complete. 

\subsection{Learning with Ontology}
\vspace{-0.08cm}
\kr{In order to support ML, 
we need to specify the input and output.
To this end, we introduce the notions of learning sample ontology (LSO) and target entailment.
We use an LSO as an input for ML methods, and the truth of a target entailment as an output.
A learning domain in ML equals to a combination of an LSO set (i.e., a dataset) and a target entailment (i.e., a prediction task).}

\vspace{-0.08cm}
\begin{defn}{\label{defn:as}}\textbf{(Learning Sample Ontology (LSO))}\\
A learning sample ontology  
$\mathcal{O} = \langle \langle \mathcal{T}, \mathcal{A} \rangle, S \rangle$
is an ontology \l\T,\A\r\ annotated by property-value pairs $S$. 
\kr{Its ABox entailment closure $\left\{g | \mathcal{T} \cup \mathcal{A} \models g \right\}$ is denoted as $\mathcal{G}(\mathcal{O})$.}
\end{defn}
\vspace{-0.12cm}

The annotation $S$ in Definition \ref{defn:as} acts as key dimensions to uniquely identify an input sample of ML methods. \kr{When the context is clear, sometimes we also use LSO to refer to its ontology  \l\T,\A\r.
By entailment reasoning with both TBox and ABox axioms,
we get a complete set of ABox entailments i.e., $\mathcal{G}(\mathcal{O})$ for modeling the input sample.}

\vspace{-0.1cm}
\begin{ex}{\label{ex:im}}\textbf{(An LSO on  Departure Flights )}\\
\kr{
Figure~\ref{fig:ontology_fig} displays some  axiom examples  of an LSO
annotated by property-value pairs $S:=$ $\{dat:01/01/2018$, $car:DL$, $ori:LAX$, $des:JFK\}$. 
The LSO corresponds to one ML input sample that is related to a flight departure from Los Angeles International Airport (LAX) to John F. Kennedy International Airport (JFK) on 01/01/2018, carried by Delta Air Lines (DL). 
The examples include some TBox axioms (1)-(6) and ABox axioms (7)-(24),
with some atomic concepts (e.g., $Airport$), 
defined concepts (e.g., $DelayedDep$),
individuals(e.g. $LAX$) and 
roles (e.g., $hasCarrier$). 
}
\vspace{-0.1cm}
\end{ex}
\vspace{-0.2cm}

\vspace{-0.3cm}
\begin{figure}[h!]
\begin{center}
\setlength{\fboxsep}{0.2cm}
\framebox{%
\begin{minipage}[t][5cm]{8.1cm}
\begin{gargantuan} 
\vspace*{-0.23cm}
\begin{equation}
\hspace*{-0.5cm} 
Dep\,\sqcap\,\exists hasDelMin.\{Pos\} \sqsubseteq DelayedDep
\label{eq:Delayed}
\end{equation}
\vspace*{-0.5cm}
\begin{equation}
\hspace*{-0.5cm} 
Dep\,\sqcap\,\exists hasDelMin.\{Neg\} \sqsubseteq OnTimeDep 
\label{eq:OnTime}
\end{equation}
\vspace*{-0.5cm}
\begin{equation}
\hspace*{-0.5cm} 
hasCarrier \circ hasCarHub \sqsubseteq hasDepHub
\label{eq:hasDepHub}
\end{equation}
\vspace*{-0.5cm}
\begin{equation}
\hspace*{-0.5cm} 
hasNebApt \circ hasRecDep \sqsubseteq hasRecNebDep
\label{eq:hasDepHub}
\end{equation}
\vspace*{-0.5cm}
\kr{
\begin{equation}
\begin{aligned}
\hspace*{-0.5cm} 
Dep \sqcap \exists hasOri.\left\{ CA \right\} \sqcap \exists hasDes.\left\{ CA \right\} 
\sqsubseteq \exists withIn.\{CA\}
\label{eq:CADep}
\end{aligned}
\end{equation}
\vspace*{-0.5cm}
\begin{equation}
\hspace*{-0.5cm} 
\exists withIn.\top \sqsubseteq InStateDep
\label{eq:instateDep}
\end{equation}
\vspace*{-0.25cm}
}
\vspace{-0.6cm}
\setlength{\columnsep}{0.3cm}
\begin{multicols}{2}\noindent
\begin{equation}
Airport(LAX)\label{eq:LAX}
\end{equation}
\begin{equation}
locatedIn(LAX, CA)\label{eq:lax_ca}
\end{equation}
\end{multicols}
\vspace*{-0.24cm}
\vspace{-0.6cm}
\setlength{\columnsep}{0.3cm}
\begin{multicols}{2}\noindent
\begin{equation}
Carrier(DL) \label{eq:dl}
\end{equation}
\begin{equation}
Departure(d)\label{eq:d1_departure}
\end{equation}
\end{multicols}
\vspace*{-0.24cm}
\vspace{-0.6cm}
\setlength{\columnsep}{0.3cm}
\begin{multicols}{2}\noindent
\begin{equation}
hasDelMin(d, Pos)\label{eq:dlp}
\end{equation}
\begin{equation}
hasWea(d, wea)\label{eq:d_wea}\\
\end{equation}
\end{multicols}
\vspace*{-0.24cm}
\vspace{-0.60cm} 
\setlength{\columnsep}{0.3cm}
\begin{multicols}{2}\noindent
\begin{equation}
hasOri(d, LAX)
\label{eq:d1_LAX}
\end{equation}
\begin{equation}
hasCarrier(d, DL)
\label{eq:d1_DL} 
\end{equation}
\end{multicols}
\vspace*{-0.85cm}
\setlength{\columnsep}{0.3cm}
\begin{multicols}{2}\noindent
\begin{equation}
Airport(JFK)\label{eq:JFK}
\end{equation}
\begin{equation}
hasDes(d, JFK)\label{eq:d_jfk}
\end{equation}
\end{multicols}
\vspace*{-0.85cm}
\vspace{0.00cm} 
\setlength{\columnsep}{0.3cm}
\begin{multicols}{2}\noindent
\begin{equation}
LAX = ori
\label{eq:LAX_ori}\\
\end{equation}
\begin{equation}
DL = car \label{eq:DL_car}
\end{equation}
\end{multicols}
\vspace*{-0.85cm}
\setlength{\columnsep}{0.3cm}
\begin{multicols}{2}\noindent
\begin{equation}
hasRecDep(d, d_{1})
\label{eq:d1_d11}\\
\end{equation}
\begin{equation}
hasCarrier(d_{1}, MU)\label{eq:d2_mu}
\end{equation}
\end{multicols}
\vspace*{-0.85cm}
\setlength{\columnsep}{0.3cm}
\begin{multicols}{2}\noindent
\begin{equation}
hasRecDep(d, d_{2})
\label{eq:d1_d12}\\
\end{equation}
\begin{equation}
hasCarrier(d_{2}, AA)\label{eq:d12_aa}
\end{equation}
\end{multicols}
\vspace*{-0.85cm}
\setlength{\columnsep}{0.3cm}
\begin{multicols}{2}\noindent
\begin{equation}
DelayedDep(d)\label{eq:d1_delay}
\end{equation}
\begin{equation}
HeavySnow(wea) \label{eq:wea}
\end{equation}
\end{multicols}
\end{gargantuan}
\end{minipage}
}
\end{center}
\vspace{-0.4cm}
\caption{\label{fig:ontology_fig}
\kr{Ontology Examples of An LSO on Departure Flights}}
\end{figure}
\vspace{-0.15cm}

Definition ~\ref{defn:mld} revisits the concept of \textit{learning domain} in ML 
and defines \textit{target entailment}.
A learning domain is also annotated by property-value pairs (cf. Definition~\ref{defn:da}).
\begin{defn}
{\label{defn:mld}}\textbf{(Learning Domain and Target Entailment)}\\
A  learning domain $\mathcal{D} = \langle  \mathbb{O}, g^t\rangle$ consists of 
a set of LSOs $\mathbb{O}$  that share the same TBox $\mathcal{T}$,
and a target entailment $g^t$ whose 
\kr{truth in an LSO
is to be predicted.}
\end{defn}

\begin{defn}{\label{defn:da}}\textbf{(Learning Domain Annotation)}\\
The annotation property-value pairs of the learning domain  $\mathcal{D}$ in Definition \ref{defn:mld}
 are defined as \kr{$S(\mathcal{D}) = 
 (\cap_{\langle \langle \mathcal{T},\mathcal{A} \rangle, S \rangle 
 \in \mathbb{O}}S) \cup \left\{t\_e: g^t\right\}$}.
\end{defn}

\begin{ex}{\label{ex:da}}\textbf{(
A Learning Domain on Departure Flights)}\\
\kr{Now we consider a learning domain \Dx 0= \l \bOx 0, $g^t_0$\r,  
where the target entailment $g^t_0$ is  $DelayedDep(d)$ and \bOx 0 contains the LSO in Example~\ref{ex:im},
as well as many similar LSOs with the same (carrier, origin airport and destination airport),
but different dates. 
The domain annotation $S$(\Dx 0) is
$\left\{car:DL, ori:LAX, des:JFK, 
t\_e: DelayedDep(d) \right\}$.
}
\end{ex}
\vspace{-0.2cm}

With the above definitions, Definition~\ref{defn:lwo} revisits the notion of \textit{within domain supervised learning} task \cite{mohri2012foundations}.
\kr{It reduces a prediction problem to a supervised learning problem with steps of learning and predicting.}

\vspace{-0.1cm}
\begin{defn}{\label{defn:lwo}}\textbf{(Within Domain Supervised Learning)}\\
\kr{Given a learning domain $\mathcal{D} = \langle \mathbb{O}, g^t \rangle$, 
whose LSOs $\mathbb{O}$ are divided into two disjoint sets $\mathbb{O}^{\prime}$ and $\mathbb{O}^{\prime\prime}$,
a supervised learning task within $\mathcal{D}$,
denoted by
$\mathcal{L} = \langle \mathcal{D}, \mathbb{O}^{\prime}, \mathbb{O}^{\prime\prime}, \mathcal{M} \rangle$, is a task of
learning a model $\mathcal{M}$ with $\mathbb{O}^{\prime}$ and $g^t$ 
to predict 
the truth of $g^t$ in each $\mathcal{O}$ in $\mathbb{O}^{\prime\prime}$.
Here, $\mathbb{O}^{\prime}$ is called a training LSO set, 
while $\mathbb{O}^{\prime\prime}$ is called a testing LSO set.
}
\end{defn}
\vspace{-0.1cm}

\kr{
In the training LSO set, we assume the ABox axioms (observations) are complete.
The target entailment is true if it is entailed by an LSO, and false otherwise.
In the testing LSO set, we assume the ABox axioms are incomplete (some observations are missing or we predict before they are observed), 
and the truth of the target entailment is predicted by the model. 
}

\vspace{-0.08cm}
\kr{
\begin{ex}{\label{ex:lwo}}\textbf{(Within Domain Supervised Learning)}\\
Given the learning domain $\mathcal{D}_0$ in Example \ref{ex:da}, 
we train the model $\mathcal{M}$ with LSOs before or at date $t_0$ (i.e., $\mathbb{O}^{\prime}$), 
and apply the model to predict the truth of $DelayedDep(d)$ in each LSO after $t_0$ (i.e., $\mathbb{O}^{\prime\prime}$). 
\end{ex}
}

\kr{
To feed an ML algorithm, 
each LSO in both training LSO set and testing LSO set is encoded
into a real value vector, denoted as $\bm{x}$.
We first transform it into (i) a value vector $\bm{v}$ with data properties by concatenating their numeric values and 
(ii) an entailment vector 
$\bm{e}$
by BOE embedding with a set of entailments \kr{entailed} by domain LSOs (cf. Definition \ref{defn:boe}).
Then we concatenate $\bm{e}$ and $\bm{v}$ as the real value vector: $\bm{x} = \left[\bm{e},\bm{v}\right]$.
The target entailment $g^t$ in each training LSO is transformed into a binary existence variable, denoted as $y$.
It's assigned $1$ if $g^t \in \mathcal{G}(\mathcal{O})$, and $0$ otherwise.
}

\vspace{-0.08cm}
\begin{defn}{\label{defn:boe}}\textbf{(Bag of Entailments)}\\
Given an entailment set $\left\{g_i |i=1,...,n \right\}$,
Bag of Entailment (BOE) is an ontology encoding method that represents an LSO (with ABox entailment closure $\mathcal{G}$) to a vector $\bm{e} = (e_1,e_2,...,e_n)$ where $e_{i}=1$ if $g_i \in \mathcal{G}$ and $0$ otherwise.
\end{defn}

\subsection{Explaining Transfer Learning}
 
\kr{Definition \ref{defn:ont_tr} revisits the concepts of \textit{transfer learning}, \textit{positive transfer} and \textit{negative transfer} \cite{pan2010survey}.
}

\vspace{-0.1cm}
\begin{defn}{\label{defn:ont_tr}}\textbf{(Transfer Learning)}\\
\kr{
Given two learning domains 
$\mathcal{D}_{\alpha}$ $=$ $\langle  \mathbb{O}_{\alpha}, g_{\alpha}^t\rangle$ and $\mathcal{D}_{\beta}$ $=$ $\langle \mathbb{O}_{\beta}, g_{\beta}^t \rangle$, 
where the LSOs of domain $\mathcal{D}_{\beta}$ are divided into two disjoint sets $\mathbb{O}_{\beta}^{\prime}$ and $\mathbb{O}_{\beta}^{\prime\prime}$,
transfer learning from $\mathcal{D}_{\alpha}$ to $\mathcal{D}_{\beta}$, 
denoted by $\mathcal{F}_{\alpha \rightarrow \beta}$ is a 
task of learning a model $\mathcal{M}_{\alpha \rightarrow \beta}$ from
$\mathbb{O}_{\alpha}$, $g_{\alpha}^t$, $\mathbb{O}_{\beta}^{\prime}$ and $g_{\beta}^t$
to predict 
the truth of $g_{\beta}^t$ in each LSO in $\mathbb{O}_{\beta}^{\prime\prime}$.
$\mathcal{F}_{\alpha \rightarrow \beta}$ is  
defined as a positive transfer if $\mathcal{M}_{\alpha \rightarrow \beta}$ outperforms $\mathcal{M}_{\beta}$ which is learned within domain $\mathcal{D}_{\beta}$ according to Definition \ref{defn:lwo},
and a negative transfer otherwise.
}
\end{defn}
\vspace{-0.27cm}

In Definition \ref{defn:ont_tr},
$\mathcal{D}_{\alpha}$ and $\mathcal{D}_{\beta}$ are called \textit{source domain} and \textit{target domain}, respectively,
while \kr{$\mathcal{O}_{\beta}^{\prime}$ and $\mathcal{O}_{\beta}^{\prime\prime}$ are the training LSO set and testing LSO set in the target domain.
In comparison with supervised learning within the target domain (Definition \ref{defn:lwo}), 
transfer learning has the same settings except that it learns the model from not only the training LSO set but also the LSO set from a source domain.

The effect of $\mathcal{F}_{\alpha \rightarrow \beta}$, namely the \textit{transferability}, 
can be measured by comparing the performance of $\mathcal{M}_{\alpha \rightarrow \beta}$ and $\mathcal{M}_{\beta}$.
In Definition \ref{defn:ont_tr}, 
positive transfer and negative are defined for qualitative description.}
We also define a metric called \textit{Feature Transferability Index} (FTI) 
\kr{(to be specified in (\ref{eq:fti}) on page \pageref{eq:fti})}
for quantitative measurement.
The higher the FTI metric, the higher the transferability.

\kr{Explaining transfer learning aims at justifying the good or bad performance of a model trained by a transfer learning algorithm with a specific parameter setting.
It describes human understandable factors that influence the transferability.}
Definition \ref{defn:ont_tr_x} defines this problem as \textit{correlation-based transfer explanation}.

\kr{
\vspace{-0.1cm}
\begin{defn}{\label{defn:ont_tr_x}}\textbf{(Correlation-based Transfer Explanation)}\\
Given a transfer $\mathcal{F}_{\alpha \rightarrow \beta}$ in Definition \ref{defn:ont_tr}, 
correlation-based transfer explanation 
is a task of inferring a set of influential factors $\mathbb{X}$ such that each $\mathcal{X}$ in $\mathbb{X}$ is correlated with $FTI$ and the absolute value of the correlation coefficient $\left\| \gamma(
FTI
, \mathcal{X}) \right\| \ge \epsilon$, 
where $\epsilon$ is a parameter in $[0,1]$.
An influential factor $\mathcal{X}$ is called an explanatory evidence.
\end{defn}
\vspace{-0.15cm}
}

In Definition \ref{defn:ont_tr_x},
the confidence of an explanatory evidence for transferability explanation is proportional to its absolute coefficient value $\left\| 
\gamma(
FTI, \mathcal{X}) \right\|$.
\kr{We abuse the notion and note it as} $\left\| \gamma(\mathcal{X}) \right\|$ in the \kr{remainder} of the paper.

Three kinds of explanatory \kr{evidence} (cf. Example \ref{ex:ote}) are proposed, 
including 
\kr{
\begin{itemize}
\item \textit{General factors} which are those statistic indexes of ABox entailments that quantify the overall knowledge variance and invariance from the source learning domain to the target learning domain,
\item \textit{Particular narrators} which are those particular ABox entailments that have a high impact on the transferability,
\item \textit{Core contexts} which are those ABox entailment combinations that have a high impact on the transferability.
\end{itemize}
}

\vspace{-0.1cm}
\begin{ex}{\label{ex:ote}}\textbf{(Explanatory \kr{Evidence})}\\
\kr{In US flight departure delay prediction, 
we consider three learning domains whose target entailments are all $DelayedDep(d)$:}
$\mathcal{D}_{(DL,ORD,LAX)}$ for Delta Airlines from ORD to LAX, 
$\mathcal{D}_{(B6,LAX,JFK)}$ for JetBlue from LAX to JFK 
and $\mathcal{D}_{(AA,ORD,SFO)}$ for American Airlines from ORD to SFO,
as well as a negative transfer from $\mathcal{D}_{(DL,ORD,LAX)}$ to $\mathcal{D}_{(B6,LAX,JFK)}$ 
and a positive transfer from domain $\mathcal{D}_{(DL,ORD,LAX)}$ to $ \mathcal{D}_{(AA,ORD,SFO)}$.
We explain the two transfers with
(i) general factors e.g., the percentage of shared entailments between $\mathcal{D}_{(DL,ORD,LAX)}$ and $\mathcal{D}_{(AA,ORD,SFO)}$ ($\mathcal{D}_{(DL,ORD,LAX)}$ and $\mathcal{D}_{(B6,LAX,JFK)}$) is high (low),
(ii) particular narrators e.g.,  entailment ``$locatedIn(ori,East)$" plays a positive role in the transfer from $\mathcal{D}_{(DL,ORD,LAX)}$ to $\mathcal{D}_{(AA,ORD,SFO)}$,
and (iii) core contexts, e.g., the entailment set composed of ``$hasOri(dep,ORD)$" and 	``$locatedIn(des,CA)$" has a high impact on the positive transfer from $\mathcal{D}_{(DL,ORD,LAX)}$ to $\mathcal{D}_{(AA,ORD,SFO)}$.
\end{ex}

\section{Method}
\subsection{\kr{Transferability Measurement}}
\kr{In transfer learning, the training of the model  for the target learning domain (i.e., $\mathcal{M}_{\alpha \rightarrow \beta}$ in Definition \ref{defn:ont_tr})
either directly integrates samples of the source learning domain 
or indirectly utilizes model parameters learned in the learning source domain \cite{pan2010survey,weiss2016survey}. 
In this study, we adopt the latter. 
Features learned by Convolutional Neural Networks (CNNs) i.e., learned parameters of hidden network layers, are transferred from the source learning domain to the target.
As transferability measurement only depends on the performance of the model trained within the target learning domain (i.e, $\mathcal{M}_{\beta}$) and the model trained with transfer (i.e., $\mathcal{M}_{\alpha \rightarrow \beta}$), 
how transfer learning is implemented does not impact the generality of our explanation framework.
}

\begin{figure}[h]
\centering
\includegraphics[scale=0.4]{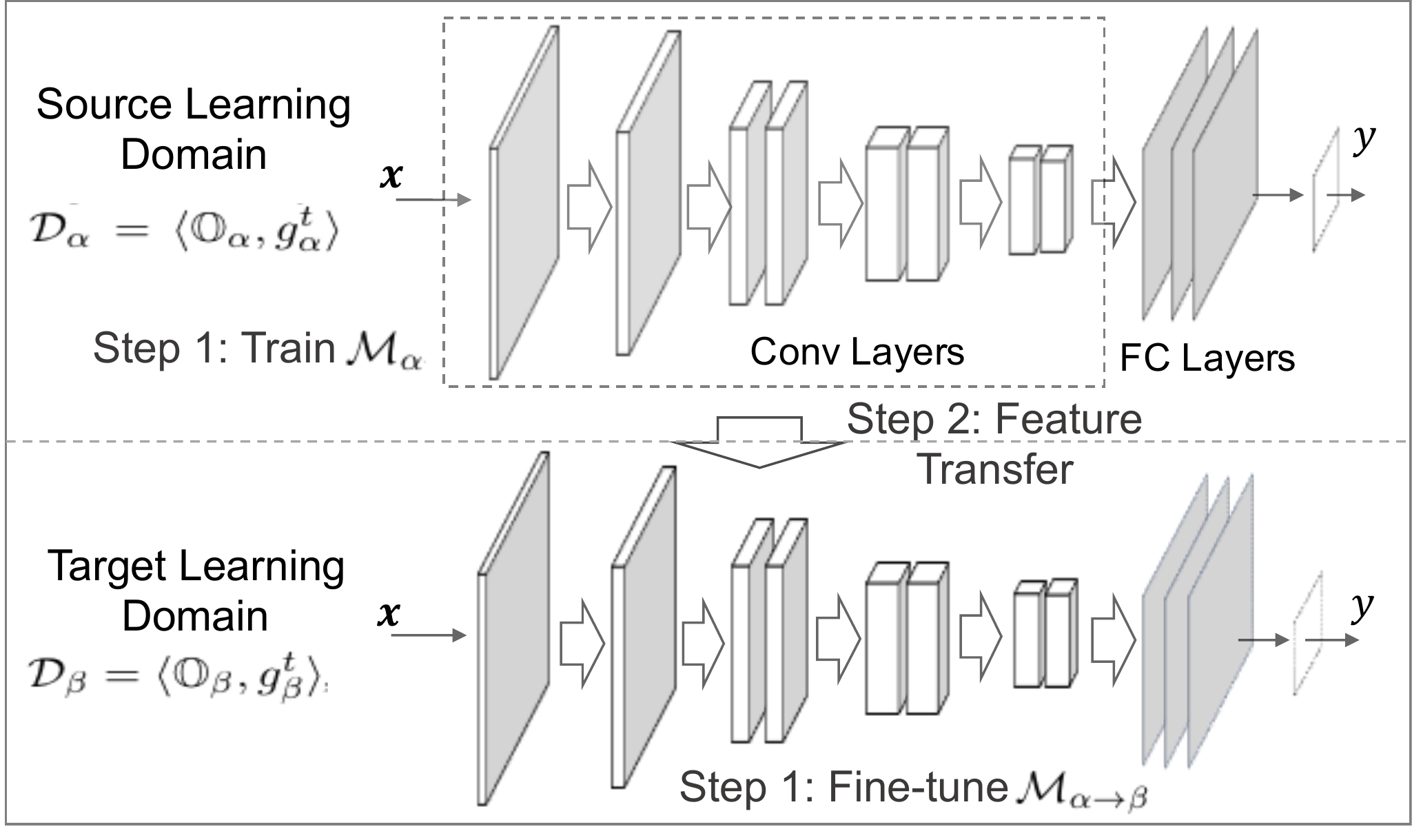}
\vspace{-0.23cm}
\caption{\kr{Transfer Learning with Convolutional Neural Networks.}
}
\label{fig:cnn}
\end{figure}

\kr{A CNN is stacked by convolutional (Conv) layers which learn the feature with data locality, and fully connected (FC) layers which learn the non-linear relationship between the input and output. 
As shown in Figure \ref{fig:cnn}, 
we first train a CNN model within the source learning domain (i.e., $\mathcal{M}_{\alpha}$) using its LSO set (i.e., $\mathbb{O}_{\alpha}$) and target entailment (i.e., $g_{\alpha}^t$), 
then transfer the model's feature (i.e., parameters of the Conv layers) to a CNN model in the target learning domain (i.e., $\mathcal{M}_{\alpha \rightarrow \beta}$) which has the same network architecture.
We eventually fine-tune the parameters of the model in the target learning domain with its training LSO set (i.e., $\mathbb{O}_{\beta}^{\prime}$) and target entailment (i.e., $g_{\beta}^t$).
It's called \textit{hard transfer} if we only fine-tune the FC layers and \textit{soft transfer} if we fine-tunes both FC layers and Conv layers.
}

\vspace{-0.2cm}

A feature's transferability depends on its \textit{specificity} to the \kr{learning domain where it is trained and its \textit{generality} \cite{yosinski2014transferable}.
By comparing the performance of the model trained within the target learning domain 
with the models trained with hard transfer and soft transfer,
the specificity and generality can be measured (cf. Definition~\ref{defn:fsi}).}

\vspace{-0.07cm}
\begin{defn}{\label{defn:fsi}}\textbf{(Feature Specificity/Generality Index)}\\
\kr{Given transfer learning in Definition \ref{defn:ont_tr},
let $\mathcal{M}_{\alpha \rightarrow \beta}^h$ and $\mathcal{M}_{\alpha \rightarrow \beta}^s$ be the models trained with hard transfer and soft transfer respectively,
Feature Specificity Index (FSI)
is defined as the performance drop of $\mathcal{M}_{\alpha \rightarrow \beta}^h$ over $\mathcal{M}_{\beta}$, 
while Feature Generality Index (FGI)
is defined as the performance gain of $\mathcal{M}_{\alpha \rightarrow \beta}^s$ over $\mathcal{M}_{\beta}$,
where the performance of all models 
are measured with the testing LSO set of the target learning domain.}
\end{defn}
\vspace{-0.05cm}

\kr{We propose a comprehensive index called \textit{Feature Transferability Index (FTI)} to measure the feature's transferability.
It is proportional to its generality and inversely proportional to its specificity (cf. Property \ref{prop:fti_p}).
The more the hard transfer or the soft transfer benefits the model in the target learning domain, the higher transferability the feature has.
With FSI and FGI, we calculate FTI as follows:
}
\begin{small}
\begin{equation}
{\label{eq:fti}}
FTI = \frac{\omega_1 \cdot FGI - \omega_2 \cdot FSI}{\omega_1 + \omega_2}
\end{equation}
\end{small}
where $\omega_1$ and $\omega_2$ are weight parameters in $[0,1]$ and are both set to $1$ in the \kr{remainder} of the paper if not specified.
\kr{We denote the FTI value from learning domain $\mathcal{D}_{\alpha}$ to learning domain $\mathcal{D}_{\beta}$ as $f_t(\mathcal{D}_{\alpha}, \mathcal{D}_{\beta})$.
}

\vspace{-0.05cm}
\begin{prop}{\label{prop:fti_p}}\textbf{(Relation between FTI and FSI/FGI)}\\
FTI is proportional to FGI and inversely proportional to FSI. 
\end{prop}

\subsection{External Knowledge}
Each learning domain is extended with \textit{external knowledge} from existent knowledge bases (KBs), such as DBPedia \cite{auer2007dbpedia}, for \kr{richer common sense knowledge about the prediction application.}
The extension includes two steps: 
(i) root individual selection,
and (ii) external knowledge matching and importing. 

\vspace{0.05cm}
\kr{
\noindent \textbf{Root Entailment and Root Individual.}
We define those entailments that play an import role in predicting the truth of the target entailment as \textit{Root Entailments}, denoted as $\mathcal{G}^R$.
Root entailments include \textit{Frequent Entailments} (cf. Definition \ref{defn:ie}) and \textit{Effective Entailments} (cf. Definition \ref{defn:ee}).
Those individuals that are involved in at least one root entailment are defined as \textit{Root Individuals}, denoted as $\mathcal{I}^R$.
}
\vspace{-0.1cm}
\begin{defn}{\label{defn:ie}}\textbf{(\kr{Frequent Entailment)}}\\
Given a learning domain $\mathcal{D} = \langle \mathbb{O}, g^t\rangle$
\kr{and its local entailment closure $\mathcal{G}(\mathbb{O}) = \cup_{\mathcal{O} \in \mathbb{O}} \mathcal{G}(\mathcal{O})$,
$g \in \mathcal{G}(\mathbb{O})$ is a frequent entailment} if
$\sfrac{\left| \left\{  \mathcal{O}  \in \mathbb{O} | g \in \mathcal{G}(\mathcal{O}) \right\} \right|}  {\left| \mathbb{O} \right|} \ge \sigma
$
, where 
$\left| \cdot \right|$ calculates the set cardinality, $\sigma$ is a parameter in $[0,1]$.
\end{defn}

\vspace{-0.15cm}
\begin{defn}{\label{defn:ee}}\textbf{(Effective Entailment)}\\
In Definition \ref{defn:ie},
a $\kappa$-element entailment subset $\mathcal{G}_{\kappa} \subseteq \mathcal{G}(\mathbb{O})$ 
is a set of effective entailments if
$r_{e} + r_{i} \ge \tau$, where 
$r_{e} = \sfrac{\left| \left\{ \mathcal{O} \in \mathbb{O} | \mathcal{G}_{\kappa} \cup \left\{ g^t \right\} \subseteq \mathcal{G}(\mathcal{O}) \right\} \right|}  {\left| \mathbb{O} \right|}$ and 
$r_{i} = \sfrac{\left| \left\{ \mathcal{O} \in \mathbb{O} | (\mathcal{G}_{\kappa} \cup \left\{ g^t \right\}) \cap \mathcal{G}(\mathcal{O}) = \emptyset \right\} \right|}  {\left| \mathbb{O} \right|}$
, $\kappa \ge 1$ and $\tau \in [0,1]$ are parameters. 
\end{defn}
\vspace{-0.1cm}

\kr{In Definition \ref{defn:ie}, we calculate the rate of LSOs that contain an entailment $g$.
The entailments that appear in a large part of LSOs are frequent entailments.
In Definition \ref{defn:ee}, 
$r_e$ ($r_i$) represents the rate of LSOs where the entailment subset $\mathcal{G}_{\kappa}$ and the target entailment $g^t$ co-exist (co-inexist).
The higher $r_e + r_i$, the more effective $\mathcal{G}_{\kappa}$ in predicting the truth of $g^t$, 
according to the theory and practice of correlation-based ML feature selection \cite{hall1999correlation}.
}

\vspace{-0.06cm}
\begin{ex}{\label{ex:ri}}\textbf{(Root Individual Selection)}\\
\kr{
We consider the learning domain $\mathcal{D}_{(B6,LAX,JFK)}$ in Example~\ref{ex:ote}, 
``$hasOri(d,LAX)$" is a frequent entailment as it appears in all the LSOs of the domain, 
while the entailment subset composed of ``$hasRecDep(d,d_2)$", ``$DelayedDep(d_2)$" and ``$hasCarrier(d_2,AA)$" are effective as they 
co-exist or co-inexist 
with the target entailment ``$DelayedDep(d)$" in a large part of LSOs.
The individuals $d$, $d_2$, $LAX$ and $AA$ that are involved in the above entailments are root individuals. 
}
\end{ex}

\kr{
\noindent \textbf{Knowledge Importing Workflow.}
For each learning domain,
we match each of its root individuals with an entity of an external KB,
and then import the concepts and roles of the entity. 
The workflow is shown in Algorithm \ref{alg:xk}.

In Line \ref{alg:line_fe1} and \ref{alg:line_match}, 
we use root individuals to match external entities by name matching.
Using root individuals significantly saves computation and storage,
as the non-root individuals take a large part but usually lead to external axioms that contribute little to the richness of explanatory \kr{evidence} (cf. Evaluation for more details).

From Line \ref{alg:line_fe2} to Line \ref{alg:line_break},
we (i) extract values of concepts and roles (i.e., $\mathcal{K}$) of each matched entity,
(ii) transform them into ABox axioms with the terminologies defined in TBox,
(iii) check their consistence with local LSOs and the constraint axioms (i.e., $\mathcal{C}$),
and (iv) add them into the set of external axioms (i.e., $\mathcal{A}_e$) (cf. Example \ref{ex:ea}).
The consistency checking is to avoid errors caused by name matching (cf. Example \ref{ex:ccea}). 
Line \ref{alg:line_infer} eventually computes the entailment closure of the learning domain, denoted as $\mathcal{G}(\mathcal{D})$, together with the local LSOs and external axioms.
}

\kr{
\vspace{-0.1cm}
\begin{ex}{\label{ex:ea}}\textbf{(External Axioms)}\\
The individual $LAX$ in Example~\ref{ex:ri} is matched with the entity $Los\_Angeles\_International\_Airport$ in DBPedia.
The triples related to the concept (``$rdf:type$")   and roles,  e.g., ``$geo:lat$" and ``$geo:long$",   ``$dbo:hubAirport$", of the entity are extracted and transformed into external ABox axioms e.g., ``$hasLat(LAX, 38.94)$". 
\end{ex}
\vspace{-0.1cm}
}

\kr{
\vspace{-0.1cm}
\begin{ex}{\label{ex:ccea}}\textbf{(Consistency Checking)}\\
In Example \ref{ex:ea},
the individual $LAX$ can be matched with the entity $L.A. International Airport$ (a song by Leanne Scott) in DBPedia by name matching.
The constraint axiom ``$Location\, \sqcap\, Song \sqsubseteq \bot $", the local axioms ``$Airport(LAX)$" and ``$Airport \sqsubseteq$ Location", and the external axiom ``$Song(LAX)$" suggest that the entity matching is incorrect.  
\end{ex}
\vspace{-0.1cm}
}

\kr{
The constraint axioms, 
which may contain concept expressions that are more expressive than DL $\mathcal{EL}^{++}$ for the requirement of a specific prediction application, 
are an extension of the TBox of the learning domain,
but are detached from the TBox to avoid increasing reasoning complexity in other steps.
}

\vspace{-0.1cm}
\begin{algorithm}[h!]
\small
\KwIn{
(i) A learning domain $\mathcal{D}=\langle g^t, \mathbb{O} \rangle$ with TBox $\mathcal{T}$,
(ii) Root individuals $\mathcal{I}^R$,
(iii) An external KB $\mathcal{B}$,
(iv) Constraint axioms $\mathcal{T}_c$
} 
\KwResult{
$\mathcal{G}(\mathcal{D})$: Entailment closure of the learning domain
}
\Begin{
$\mathcal{A}_e := \emptyset$;
\emph{$\%$ Init. of the external axiom set 
}\\
\ForEach{root individual $i \in \mathcal{I}^R$\label{alg:line_fe1}}{ 
$\mathcal{N}^i \leftarrow (\mathcal{B},i)$ \emph{$\,\,\%$ Match external entity by name.}\label{alg:line_match} \\
\ForEach{entity $e \in \mathcal{N}^i$\label{alg:line_fe2}}{ 
$\mathcal{V} \leftarrow (\mathcal{B},e)$\emph{$\,\,\%$ Extract concpepts and roles}\\
$\mathcal{K} \leftarrow (\mathcal{V},\mathcal{T})$\emph{$\,\,\%$Transform to external axioms}\\
\emph{$\,\,\%$    Consistency checking} \\
\If{ $ \mathcal{O} \cup \mathcal{T}_c \cup \mathcal{K} \not\models \bot$ for $\forall \mathcal{O} \in \mathbb{O}$\label{alg:line_if}}
{ 
$\mathcal{A}_e := \mathcal{A}_e \cup \mathcal{K}$ \\
\textbf{break} \emph{$\,\,\%$ Adopt the first matched entity}\label{alg:line_break}\\
}
}}
$\mathcal{G}(\mathcal{D}) \leftarrow \cup_{\mathcal{O} \in \mathbb{O}} 
\mathcal{G}(\mathcal{O} \cup \mathcal{A}_e)
$ \emph{$\,\,\%$ Entailment reasoning}\label{alg:line_infer}\\
}
\Return{$\mathcal{G}$($\mathcal{D}$)};
\caption{ {\label{alg:xk}} \small{{\tt{ExternalAxiomsImport}}$\langle \mathcal{D}, \mathcal{I}^R, \mathcal{B}, \mathcal{C} \rangle$}}
\end{algorithm}
\vspace{-0.2cm}

\subsection{Correlative Reasoning}
\kr{
We propose a method called correlative reasoning for calculating the explanatory evidence (i.e., general factors, particular narrators and core contexts) with entailment closures of the learning domains and the FTIs between learning domains. 
It is composed of two steps: 
evidence embedding and correlation analysis.
}

\kr{
\noindent \textbf{Evidence Embedding.}
It represents an explanatory evidence by a real value, 
without losing the evidence's semantics in analyzing the feature transferability.
Given an evidence $\mathcal{X}$, we denote its embedding as $f_e(\mathcal{X})$. 

General factors are statistic indexes that 
measure the overall difference and similarity 
between two learning domains.
Definition~\ref{defn:edcr} defines the embedding approach for three general factors: $d^{new}$, $d^{obs}$ and $d^{inv}$,
which are directed domain change rates from the source learning domain to the target, 
with \textit{new}, \textit{obsolete} and \textit{invariant} entailments respectively.
}

\kr{
\vspace{-0.1cm}
\begin{defn}{\label{defn:edcr}}\textbf{(Entailment-based Domain Change Rates)}\\
Given source learning domain $\mathcal{D}_{\alpha}$ 
and target learning domain $\mathcal{D}_{\beta}$ in transfer learning  (Definition \ref{defn:ont_tr}),
the entailment-based domain change rates from $\mathcal{D}_{\alpha}$ to $\mathcal{D}_{\beta}$ are defined as:
\vspace{-0.15cm}
\begin{equation}\label{eq:ebdcr}
\vspace{-0.05cm}
\begin{cases}
d^{new} = \frac{\left| \left\{ g | g \in \mathcal{G}(\mathcal{D}_{\beta}), g \not\in \mathcal{G}(\mathcal{D}_{\alpha})  \right\}  \right|}{\left| \mathcal{G}(\mathcal{D}_{\beta}) \right|}  \\
d^{obs} = \frac{\left| \left\{ g | g \in \mathcal{G}(\mathcal{D}_{\alpha}), g \not\in \mathcal{G}(\mathcal{D}_{\beta})  \right\}  \right|}{\left| \mathcal{G}(\mathcal{D}_{\alpha}) \right|}  \\
d^{inv} = \frac{\left| \left\{ g | g \in \mathcal{G}(\mathcal{D}_{\alpha}), g \in \mathcal{G}(\mathcal{D}_{\beta})  \right\}  \right|}{\left| \mathcal{G}(\mathcal{D}_{\alpha}) \cup \mathcal{G}(\mathcal{D}_{\beta}) \right|}  \\
\end{cases}
\vspace{-0.05cm}
\end{equation}
where the operation $\left| \cdot \right|$ calculates set cardinality.
\end{defn}
\vspace{-0.1cm}
}

\kr{
\vspace{-0.1cm}
\begin{ex}{\label{ex:edcr}}\textbf{(Entailment-based Domain Change Rates)}\\
In the transfer from domain $\mathcal{D}_{(DL,ORD,LAX)}$ (entailment closure size: $25180$) to
domain $\mathcal{D}_{(B6,LAX,JFK)}$ (entailment closure size: $13412$) in Example \ref{ex:ote},
the sizes of new, obsolete and invariant entailments are $11419$, $23187$ and $1193$. 
Thus the domain change rates $d^{new}$, $d^{obs}$ and $d^{inv}$ are calculated as $\frac{11419}{13412}$, $\frac{23189}{25180}$ and $\frac{1193}{13412+25180}$ respectively.
\end{ex}
\vspace{-0.1cm}
}

\kr{
A particular narrator is one single entailment that (i) is shared by the source and target learning domains, and (ii) has positive or negative impact on the feature's transferability.
Core context is an extension of particular narrator from one single entailment to an entailment set (combination). 
To simplify the representation, we regard a particular narrator as a one-element entailment set,
and use evidence note $\mathcal{X}$ to denote the entailment set involved in a particular narrator or core context.

Definition \ref{defn:dece} defines the embedding approach for particular narrators and core contexts.
It transforms a specific entailment or an entailment set into a binary variable called DEC with the entailments' co-existence in the source and target learning domains considered.
}

\kr{
\vspace{-0.1cm}
\begin{defn}{\label{defn:dece}}\textbf{(Directed Entailment Co-existence)}\\
Given source learning domain $\mathcal{D}_{\alpha}$ 
and target learning domain $\mathcal{D}_{\beta}$ in transfer learning  (Definition \ref{defn:ont_tr}),
Directed Entailment Co-existence (DEC) of an entailment set $\mathcal{G} \subseteq \mathcal{G}(\mathcal{D}_{\alpha})$ 
is $1$ if $\mathcal{G} \subseteq \mathcal{G}(\mathcal{D}_{\beta})$ 
and $0$ otherwise. 
When $\left| \mathcal{G} \right| = 1$, it calculates the DEC of a single entailment.
\end{defn}
\vspace{-0.1cm}
}

\kr{
\vspace{-0.1cm}
\begin{ex}{\label{ex:dece}}\textbf{(Directed Entailment Co-existence)}\\
In the transfer from learning domain $\mathcal{D}_{(DL,ORD,LAX)}$ to
learning domain $\mathcal{D}_{(AA,ORD,SFO)}$ in Example \ref{ex:ote},
the particular narrator of ``$hasOri(d,ORD)$" is embedded into $1$, 
while the core context composed of ``$hasOri(d,ORD)$" and ``$hasCarrier(d,DL)$" is embedded into $0$. 
\end{ex}
\vspace{-0.1cm}
}

\kr{
Different from BOE embedding in Definition \ref{defn:boe},
which transforms entailments of an LSO into one vector as ML input, 
evidence embedding transforms the entailment-based change from one learning domain to another into a real value for transferability analysis. 
}

\kr{
\noindent \textbf{Correlation Analysis.}
We analyze the correlation between a given explanatory evidence $\mathcal{X}$ and its impact on the transferability of a feature 
with a set of learning domains $\mathbb{D}$,
as shown in Algorithm \ref{alg:ca}.
Line \ref{alg:ca_fe} traverses each pair of source and target learning domains, 
which correspond to one transfer. 
For particular narrators and core contexts,
Line \ref{alg:ca_if} and \ref{alg:ca_continue} skip  the transfers whose source learning domains fail to entail all the entailments that are involved the evidence.
Line \ref{alg:ca_ee} and \ref{alg:ca_fti} calculate the embedding of the evidence and the FTI value of the transfer, respectively.
Line \ref{alg:ca_corr} calculates the Pearson Correlation Coefficient \cite{lee1988thirteen} and its p-value in a t-test with non-correlation hypothesis.
The algorithm eventually returns the correlation coefficient (cf. Definition~\ref{defn:ont_tr_x}) and its p-value.
}

\kr{
An evidence $\mathcal{X}$ is a valid explanatory evidence for the feature transferability if  
(i) its absolute value of correlation coefficient $\left\| \gamma(\mathcal{X}) \right\| \geq \epsilon$ (cf. Definition \ref{defn:ont_tr_x})
and (ii) the correlation analysis is significant (i.e., $\rho(\mathcal{X}) \leq 0.05$). 
}

\kr{
In correlative reasoning, $\sfrac{m(m-1)}{2}$ times FTI calculation are totally needed,  
where $m$ is the size of the given domain set $\mathbb{D}$.
Meanwhile, correlative reasoning costs $\sfrac{m(m-1)}{2}$ times evidence embedding calculation and one time correlation analysis calculation for each evidence. 
On the other hand, to compute a complete set of explanatory \kr{evidence}, we need to traverse all the candidate \kr{evidence}.
The number of general factors is a constant, 
while the numbers of particular narrators and core contexts are $n$ and $2^n-1$ respectively, where $n$ is the size of the entailment closure of the given learning domains (i.e., $\cup_{\mathcal{D}\in\mathbb{D}}\mathcal{G}(\mathcal{D})$).
Directly searching all the core contexts is impractical; thus we need some optimized methods of searching for core context .
}

\begin{algorithm}[h!]
\small
\KwIn{
(i) A set of learning domains $\mathbb{D} = \left\{ \mathcal{D}_k | k=1,..., m\right\}$ and
(ii) an explanatory evidence $\mathcal{X}$
} 
\KwResult{
$\mathbb{}$Correlation coefficient $\gamma(\mathcal{X})$ and p-value $\rho(\mathcal{X})$ 
}
\Begin{
$\bm{v}_{e} := \emptyset$ \emph{$\,\,\%$ Init. of evidence value array }\label{alg:ca_init1}\\
$\bm{v}_{f} := \emptyset$ \emph{$\,\,\%$ Init. of FTI value array } \label{alg:ca_init2}\\
\ForEach{ $\mathcal{D}_{k_1} \in \mathbb{D}$, $\mathcal{D}_{k_2} \in \mathbb{D}$ such that $\mathcal{D}_{k_1} \ne \mathcal{D}_{k_2}$
\label{alg:ca_fe}}{
\If{($\mathcal{X}$ is a particular narrator or a core context) and ($\mathcal{X} \not \subseteq \mathcal{G}(\mathcal{D}_{k_1})$)\label{alg:ca_if}}{
\textbf{Continue} \label{alg:ca_continue}\\
}
\emph{$\%$ Cal. evidence embedding} \\
$f_e(\mathcal{X}) \xleftarrow{Def. \ref{defn:edcr}, Def.\ref{defn:dece}} (\mathcal{X}, \mathcal{G}(\mathcal{D}_{k_1}), \mathcal{G}(\mathcal{D}_{k_2}))$ \label{alg:ca_ee} \\
$\bm{v}_{e} := [\bm{v}_{e}, f_e(\mathcal{X})]$ \\
\emph{$\%$ Cal. FTI for the transfer from $\mathcal{D}_{k_1}$ to $\mathcal{D}_{k_2}$} \\
$f_t(\mathcal{D}_{k_1}, \mathcal{D}_{k_2}) \xleftarrow{\eqref{eq:fti},Def.\ref{defn:fsi}} (\mathcal{D}_{k_1}, \mathcal{D}_{k_2})$ \label{alg:ca_fti}\\
$\bm{v}_{f} := [\bm{v}_{f}, f_t(\mathcal{D}_{k_1}, \mathcal{D}_{k_2})]$ \label{alg:ca_corr}
}
\emph{$\%$ Pearson correlation analysis and t-test} \\
$\gamma(\mathcal{X}), \rho(\mathcal{X}) = corr(\bm{v}_{e}, \bm{v}_{f})$ \label{alg:corr2}\\
}
\Return{$\gamma(\mathcal{X})$,$\rho(\mathcal{X})$};
\caption{ {\label{alg:ca}} \small{{\tt{CorrelativeReason}}$\langle \mathbb{D}, 
\mathcal{X} \rangle$}}
\end{algorithm}

\kr{
\subsection{Optimized Core Context Searching}
}
\kr{
A core context is composed of a subset of entailments of the given learning domain set 
(i.e., $\mathcal{X} \subseteq \cup_{\mathcal{D}\in\mathbb{D}}\mathcal{G}(\mathcal{D})$).
Algorithm \ref{alg:ccs} presents our core context searching algorithm. 
It starts by traversing core contexts composed of two entailments (cf. Line \ref{alg:ccs_if1} to \ref{alg:ccs_ccs1}),
and then traverses core contexts with higher dimension by adding an entailment to the current core context (cf. Line \ref{alg:ccs_fe2} to \ref{alg:ccs_ccs2}).
It adopts two approaches, {\tt EarlyStop} and {\tt FastExtend} to accelerate the search.
}

\begin{algorithm}[h!]
\small
\KwIn{
(i) A learning domain set $\mathbb{D}$,
(ii) A candidate core context evidence $\mathcal{X}$
} 
\KwResult{
Records of (evidence, coefficient, p-value)
}
\Begin{
$\mathcal{G}_0 := \cup_{\mathcal{D} \in \mathbb{D}}\mathcal{G}(\mathcal{D})$ \emph{$\,\,\%$ Cal. entailment closure} \\

\If{ $\mathcal{X} = \emptyset$ \label{alg:ccs_if1}}{

\emph{$\%$ Traverse core contexts with two entailments} \\
\ForEach{$g_1 \in \mathcal{G}_0$, $g_2 \in \mathcal{G}_0$ such that $g_1 \neq g_2$\label{alg:ccs_fe1}}{

{\tt CoreContextSearch}($\mathbb{D},\left\{g_1,g_2 \right\}$) \label{alg:ccs_ccs1}\\
}
}
\Else{
$\gamma(\mathcal{X}),\rho(\mathcal{X}) \leftarrow$ {\tt CorrelativeReason}($\mathbb{D},\mathcal{X}$) \label{alg:ccs_ca}\\
{\tt print} $\mathcal{X},\gamma(\mathcal{X}),\rho(\mathcal{X})$ \\

\If{$!$ {\tt EarlyStop} ($\mathbb{D}$,$\mathcal{X}$) 
\label{alg:ccs_if2}}{
\emph{$\%$ Traverse core contexts with one more ent.} \\

\ForEach{$g \in \mathcal{G}_0$ such that $g \not \in \mathcal{X}$\label{alg:ccs_fe2}}{

\If{\tt{FastExtend}($\mathbb{D}, \mathcal{X}, g$)}{
{\tt print} $\mathcal{X}\cup\left\{g\right\},\gamma(\mathcal{X}),\rho(\mathcal{X})$\\
}
\Else{
\tt{CoreContextSearch}($\mathbb{D}, \mathcal{X} \cup \left\{ g \right\}$) \label{alg:ccs_ccs2}\\
}

}
}
}
}
%
\caption{ {\label{alg:ccs}} \small{{\tt{CoreContextSearch}}$\langle \mathbb{D}, \mathcal{X} \rangle$}}
\end{algorithm}

\kr{
\noindent \textbf{Early Stop.}
The function {\tt EarlyStop} returns true if adding more entailments to a core context evidence will not lead to any valid core contexts,
and false otherwise.
According to Algorithm \ref{alg:ca} and the principle of t-test,
enough \textit{Evidence Domains} (cf. Definition \ref{defn:ccd}) are needed for significant correlation analysis of an evidence,
while Property \ref{prop:ccd} shows that
when a core context evidence is extended by an entailment,
the number of its evidence domains decreases. 
Thus when the correlation analysis of the current evidence is insignificant (i.e., $\rho(\mathcal{X})$ 
$>0.05$),
we stop extending this evidence with more entailments.
}

\kr{
\vspace{-0.05cm}
\begin{defn}{\label{defn:ccd}}\textbf{(Evidence Domains)}\\
Given a core context or particular narrator evidence $\mathcal{X}$ and a learning domain set $\mathbb{D}$,
the evidence domains of $\mathcal{X}$, denoted as $\mathbb{D}(\mathcal{X})$, are defined as $\left\{\mathcal{D} \in \mathbb{D} | \mathcal{X} \subseteq \mathcal{G}(\mathcal{D}) \right\}$.
\end{defn}
}

\kr{
\vspace{-0.1cm}
\begin{prop}{\label{prop:ccd}}\textbf{(Monotonicity of Evidence Domains)}\\
In Definition \ref{defn:ccd},
for all the entailment $g$ in $\cup_{\mathcal{D}\in\mathbb{D}}\mathcal{G}(\mathcal{D})$ and $g \not \in \mathcal{X}$,
we have $\mathbb{D}(\mathcal{X} \cup \left\{ g \right\}) \subseteq \mathbb{D}(\mathcal{X})$. 
\end{prop}
\vspace{-0.1cm}
}

\kr{
\noindent \textbf{Fast Extend.}
The function {\tt FastExtend} returns true if an entailment is a \textit{Synchronized Entailment} (cf. Definition \ref{defn:ce}) of another entailment in the current core context evidence, and false otherwise.
According to Lemma \ref{lem:cecc}, the new core context evidence with the synchronized entailment added has the same impact on the feature transferability as the original one.
{\tt FastExtend} enables us to directly extend a core context \kr{evidence} and avoid the calculation of evidence embedding and correlation analysis.
}

\kr{
\vspace{-0.1cm}
\begin{defn}{\label{defn:ce}}\textbf{(Synchronized Entailments)}\\
\vspace{-0.08cm}
Given a learning domain set $\mathbb{D}$,
two entailments $g_1$ and $g_2$ are synchronized, denoted by $g_1 
\overset{\underset{\mathrm{ \mathbb{D} }}{}}{=} 
g_2$, if for all the learning domain $\mathcal{D}$ in $\mathbb{D}$, $\left\{ g_1,g_2\right\} \subseteq \mathcal{G}(\mathbb{D})$ or $\left\{ g_1,g_2\right\} \cap \mathcal{G}(\mathbb{D}) = \emptyset$.  
\end{defn}
}

\kr{
\vspace{-0.1cm}
\begin{ex}{\label{ex:dece}}\textbf{(Synchronized Entailments)}\\
In our departure flights example, ``$locatedIn(LAX,LA)$" and ``$serveCity(LAX, LA)$" are synchronized entailments, w.r.t. the $92$ learning domains used in our evaluation.
\end{ex}
\vspace{-0.1cm}
}

\kr{
\vspace{-0.1cm}
\begin{lemma}{\label{lem:cecc}}\textbf{(Synchronized Evidence Extension)}\\
In Definition \ref{defn:ce}, 
given a core context or particular narrator evidence $\mathcal{X}$,
for all the entailment $g$ in $\cup_{\mathcal{D}\in\mathbb{D}}\mathcal{G}(\mathcal{D})$ and $g \not \in \mathcal{X}$,
the new core context evidence $ \mathcal{X}^{\prime} := \mathcal{X} \cup \left\{ g \right\} $ has the same correlation analysis result as $\mathcal{X}$, i.e.,
$\gamma( \mathcal{X}^{\prime} ) = \gamma(\mathcal{X}) $ 
and $\rho( \mathcal{X}^{\prime} ) = \rho(\mathcal{X}) $,
if there is an entailment $g_0$ in  
\vspace{-0.08cm}
$\mathcal{X}$ such that $g \overset{\underset{\mathrm{ \mathbb{D} }}{}}{=} g_0$. 
\end{lemma}
\vspace{-0.2cm}

\begin{proof}
$g \overset{\underset{\mathrm{ \mathbb{D} }}{}}{=} g_0$ implies $\mathbb{D}(\left\{ g \right\}) = \mathbb{D}(\left\{ g_0\right\})$ (Definition \ref{defn:ccd} and \ref{defn:ce}),
while $\left\{ g_0 \right\} \subseteq \mathcal{X}$ implies $\mathbb{D}(\mathcal{X}) \subseteq \mathbb{D}(\left\{ g_0\right\}) $ (Property \ref{prop:ccd}),
thus $\mathbb{D}(\mathcal{X}) \subseteq \mathbb{D}(\left\{ g \right\}$);
$\mathcal{X}^{\prime} = \left\{ g \right\} \cup \mathcal{X}$ implies $\mathbb{D}(\mathcal{X}^{\prime}) = \mathbb{D}(\mathcal{X}) \cap \mathbb{D}(\left\{ g \right\})$ (Definition \ref{defn:ccd}); 
thus $\mathbb{D}(\mathcal{X}^{\prime}) = \mathbb{D}(\mathcal{X})$;
\vspace{-0.08cm}
thus $\mathcal{X}$ and $\mathcal{X}^{\prime}$ have the same FTI values ($\bm{v}_{f}$) in Algorithm \ref{alg:ca}.
Meanwhile, $g_0 \in \mathcal{X}$ and $g \overset{\underset{\mathrm{ \mathbb{D} }}{}}{=} g_0$ imply 
$f_e(\mathcal{X}) = f_e(\mathcal{X}^{\prime})$ in Algorithm \ref{alg:ca} (Definition \ref{defn:dece} and \ref{defn:ce});
Thus $\mathbb{D}(\mathcal{X}^{\prime}) = \mathbb{D}(\mathcal{X})$ implies 
$\mathcal{X}$ and $\mathcal{X}^{\prime}$ have the same evidence embedding ($\bm{v}_{e}$) in Algorithm \ref{alg:ca}.
$\mathcal{X}$ and $\mathcal{X}^{\prime}$ have the same $\bm{v}_{e}$ and $\bm{v}_{f}$ imply $\gamma( \mathcal{X}^{\prime} ) = \gamma(\mathcal{X}) $ and $\rho( \mathcal{X}^{\prime} ) = \rho(\mathcal{X}) $.
\end{proof}
\vspace{-0.1cm}
}

\kr{Synchronized entailments are common especially when a large number of external axioms are imported. 
The time complexity of computing all the synchronized entailment pairs is $O(\sfrac{n(n-1)}{2})$, where $n$ is the size of $\cup_{\mathcal{D} \in \mathbb{D}}\mathcal{G}(\mathcal{D})$.
Meanwhile,
with the transitivity property of synchronized entailment (cf. Property \ref{prop:tc}),
we can merge two sets of synchronized entailments if an entailment of one set is synchronized with an entailment of another set, thus quickly calculating clusters of synchronized entailment.
}

\kr{
\begin{prop}{\label{prop:tc}}\textbf{(Transitivity of Synchronized Entailment)}\\
In Definition ~\ref{defn:ce}, $(g_1 \overset{\underset{\mathrm{ \mathbb{D} }}{}}{=} g_2)  \land (g_2 \overset{\underset{\mathrm{ \mathbb{D} }}{}}{=} g_3)
\rightarrow (g_1 \overset{\underset{\mathrm{ \mathbb{D} }}{}}{=} g_3)$.
\end{prop}
}

\kr{
Heuristics can also be developed to approximately search the core contexts.
For example, in extending a core context, 
we can either ignore entailments that are not particular narrators, 
or only add entailments that are semantically close to the core context (e.g., about the same carrier).  
They are left in our future work.
}

\section{Evaluation}\label{sec:eva}
\noindent \textbf{Experiment Setting.}
In the experiment, 
we predict whether a flight's departure will be delayed or not, with observations of recent and surrounding flights, as well as meteorology\footnote{\scriptsize{ Codes and data: https://github.com/ChenJiaoyan/X-TL}}. 
\kr{
The target entailment is set to $DelayedDep(d)$ for all the learning domains,
and carrier, origin airport and destination airport are used to identify a learning domain.}
92 learning domains composed of 10 airports and 11 carriers in US are adopted.
One learning domain has 1,880 to 9,500 LSOs extracted from 01/01/2010 to 07/01/2017. 
8372 transfers are evaluated, where FTI is measured with Area Under ROC Curve, a widely used performance metric for the prediction model. 
\kr{In deciding a valid evidence, the coefficient threshold $\epsilon$ in Definition \ref{defn:ont_tr_x} is set to $0.1$.
}

\begin{figure}[h]
\centering
\includegraphics[scale=0.49]{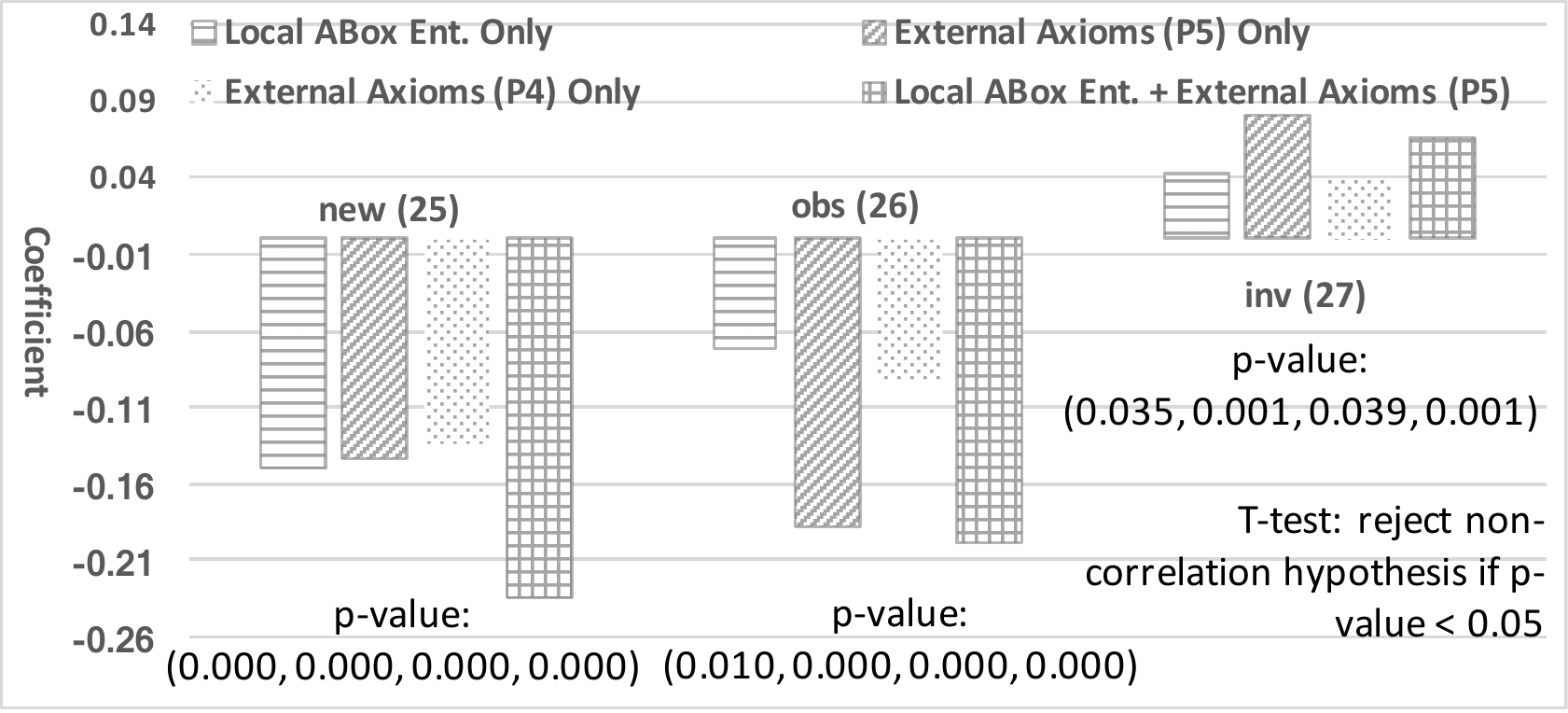}
\vspace{-0.6cm}
\caption{
General Factors $d^{new}$, $d^{obs}$ and $d^{inv}$ Calculated with Different Knowledge Parts, and Parameter Settings of P4 and P5. 
}
\label{res:factor}
\end{figure}

\begin{figure}[h]
\centering
\includegraphics[scale=0.51]{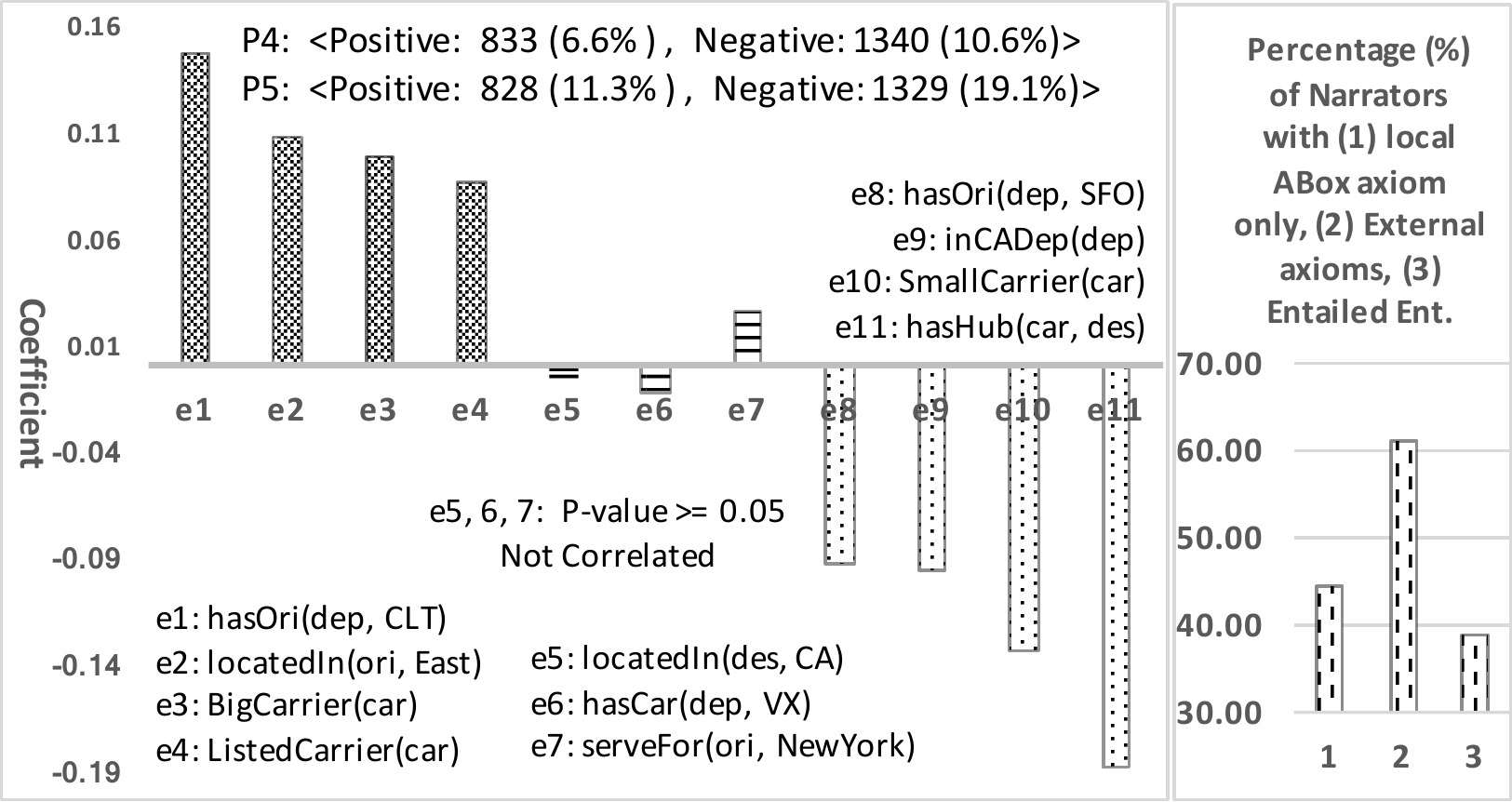}
\vspace{-0.6cm}
\caption{\kr{Examples and Statistics of Particular Narrators.}}
\label{res:narrator}
\end{figure}

\begin{figure}[h]
\centering
\includegraphics[scale=0.44]{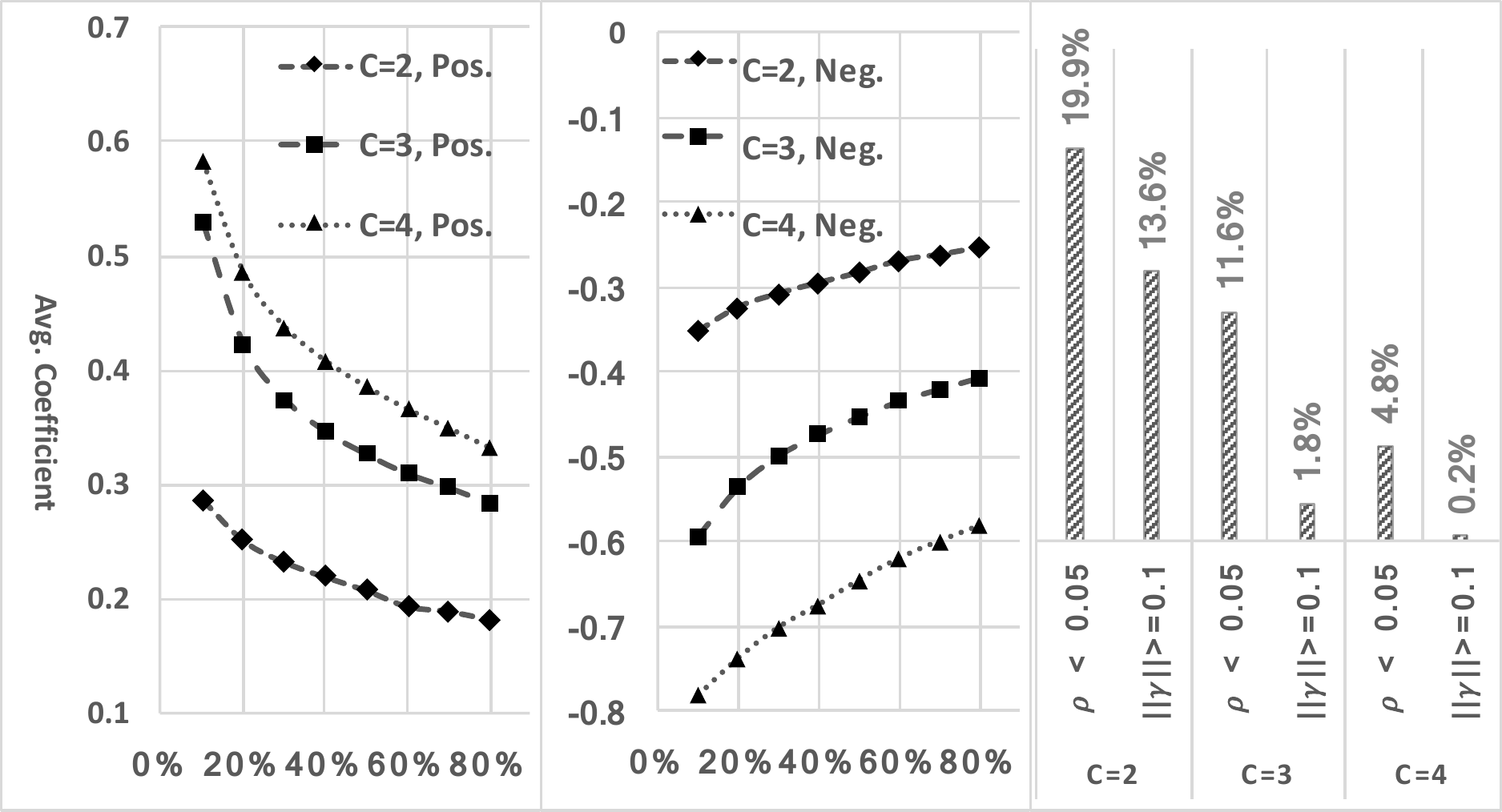}
\vspace{-0.55cm}
\caption{{[Left]}
({[Middle]}) Average Correlation Coefficient of Top $k\%$ Most Positively (Negatively) Correlated Core Contexts,
{[Right]} Percentage of Valid Core Contexts 
(i.e., $\rho(\mathcal{X})<0.05$ and $\left\| \gamma(\mathcal{X}) \right\| \geq 0.1$) 
, with Dimension $C=2,3,4$.}
\label{res:core}
\end{figure}

We report results of (i) average number of root entailments, root individuals and external axioms per learning domain (Table~\ref{res:count}),
(ii) general factors (Figure~\ref{res:factor}), 
(iii) particular narrators (Figure~\ref{res:narrator}) 
and (iv) core contexts (Figure~\ref{res:core}),
\kr{and at the same time analyze the impact of entailment reasoning and external knowledge importing on the explanation.}

\noindent \textbf{External Knowledge.}
Table~\ref{res:count} presents that the size of \kr{root entailments (including root concept assertion entailments and role assertion entailments)},
root individuals and external axioms all decreases when the parameters ($\sigma$, $\kappa$, $\tau$) increase from P1 to P5.
When ($\sigma$, $\kappa$, $\tau$) are set to P5,
only 9.3\% of the individuals are selected as root individuals, 
reducing external axioms from around $21,000$ to $615$.

On the other hand, 
importing less external axioms by selecting root individuals does not harm \kr{the richness of explanatory evidence}.
Firstly, Figure~\ref{res:factor} (page~\pageref{res:factor}, more explanatoins below) reports that setting ($\sigma$, $\kappa$, $\tau$) to P4
($6271$ external axioms imported) does not help infer more confident general factors than 
P5 ($615$ external axioms imported).
\kr{In contrast, it reduces the confidence of general factors
$d^{new}$, $d^{obs}$ and $d^{inv}$
by ($7.5\%$, $51.3\%$, $52.4\%$), when they are measured with external axioms alone.
It means those additional external axioms in setting P4 bring more noise than effective information to general factors in explaining the transferability.}
Secondly,  Figure~\ref{res:narrator} [Left] (page\pageref{res:narrator}) reports that the richness of particular narrators is kept from setting P4 to P5, 
as the total number decreases very little (e.g., positive entailments decrease from 833 to 828).

\vspace{-0.2cm}
\begin{table}[h!]
\begin{smallermathTable}
\scriptsize{
\centering
\begin{tabular}[t]{p{1.52cm}<{\centering}|p{1.3cm}<{\centering}|p{1.3cm}<{\centering}|p{1.3cm}<{\centering}|p{0.98cm}<{\centering}}
\hline
TBox Axi.: 541 & Concept Ast. Ent.: 1824 &Role Ast. Ent.: 4528 & Individual: 1159 & Ext. Axi.: $\sim21$ K \\\hline 
\hline
Parameters ($\sigma, \kappa, \tau$) & Root Concept Ast. Ent.& Root Role Ast. Ent. & Root Individual & External Axioms\\\hline
P1:($.90, 1, .40$) & $1105$ ($61\%$)  & $3805$ ($84\%$) & $1103$ ($95\%$) & $\sim$ $20$ K \\\hline
P2:($.93, 1, .43$) & $990$ ($54\%$)  & $3459$ ($76\%$) & $1080$ ($93\%$) & $\sim$ $19$ K \\\hline
P3:($.96, 1, .46$) & $540$ ($30\%$)  &  $1816$ ($40\%$)& $872$ ($75\%$) & $\sim$ $16$ K \\\hline
P4:($.99, 1, .49$) & $305$ ($17\%$)  &$980$ ($22\%$)  &$510$ ($44\%$)  & $6271$ \\\hline
P5:($.99, 2, .49$) & $157$ ($8.6\%$)  &  $402$ ($8.9\%$)& $108$ ($9.3\%$)  & $615$ \\\hline 
\end{tabular}
\vspace{-0.25cm}
\caption{\label{res:count} Average Number of Root Entailments, Root Individuals and External Axioms per Learning Domain.}
}
\end{smallermathTable}
\end{table}
\vspace{-0.2cm}

\vspace{0.07cm}
\noindent \textbf{General Factors.}
Figure~\ref{res:factor} (\kr{Local ABox Ent. + External Axioms (P5)}) presents that 
general factors $d^{obs}$ and $d^{new}$ have a significant negative impact on the feature's transferability ($\gamma(\mathcal{X}) < -0.2$).
Thus we can explain a negative transfer like $\mathcal{F}_{(DL,ORD,LAX)\rightarrow(B6,LAX,JFK)}$ with explanations like \textit{``There are a high percentage of new and obsolete entailments from domain $\mathcal{D}_{(DL,ORD,LAX)}$ to $\mathcal{D}_{(B6,LAX,JFK)}$"}.
On the other hand,
there is positive correlation between general factor $d^{inv}$ and FTI ($\rho(\mathcal{X}) < 0.05$), 
which is opposite to $d^{obs}$ and $d^{new}$,
but the correlation is weak ($\left\| \gamma(\mathcal{X}) \right\|< 0.08$).
Therefore, sharing a large percentage of entailments is not a confident evidence to explain a positive feature transfer.

\vspace{-0.2cm}
\vspace{-0.3cm}

\noindent \textbf{Entailment Narrator.} 
Figure \ref{res:narrator} [Left] shows that 11.3\% (19.1\%) of \kr{the entailments are positively (negatively) correlated with FTI in parameter setting P5}.
Those entailments are adopted as particular narrators for explaining a positive or negative feature transfer.
According to the particular narrator examples, 
we can explain the positive transfer $\mathcal{F}_{(DL,ORD,LAX) \rightarrow (AA,ORD,SFO)}$ with descriptions like 
(i) \textit{``the origin airport of both source and target learning domains is in the east part of US" (e2)} and (ii) \textit{``the carriers of both source and target learning domains are public companies" (e4)}.
We can explain the negative transfer $\mathcal{F}_{(DL,ORD,LAX)\rightarrow(B6,LAX,JFK)}$ with descriptions like
\textit{``the carriers of both source and target learning domains are small companies; it's hard to transfer a feature between two learning domains with small carriers" (e10)}.

\vspace{-0.1cm}
\vspace{-0.1cm}

\vspace{0.08cm}
\noindent \textbf{\kr{Entailment Reasoning} and External Knowledge.} 
Figure~\ref{res:factor} \kr{(Local ABox Ent. Only vs. External Axioms (P5) vs. Local ABox Ent. + External Axioms (P5))} shows that 
combing local ABox entailments and external axioms for $d^{obs}$ ($d^{new}$) achieves $178.9\%$ ($56.0\%$) and $5.9\%$ ($62.5\%$)
higher absolute coefficient than using local ABox entailments alone
and using external axioms alone respectively.
This verifies external knowledge's positive impact on the confidence of general factors.
Meanwhile, Figure~\ref{res:narrator} [Right] shows that \kr{44.4\% of particular narrators use local ABox axioms only, 
while 61.1\% and 38.9\% of them involve external axioms and \kr{entailed} entailments respectively}.
This verifies the positive impact of \kr{entailment reasoning} and external knowledge on the quality of particular narrators. 

\noindent \textbf{Core Context.}
Figure~\ref{res:core} [Left] and [Middle] present that the core contexts composed of 2 to 4 entailments \kr{have much higher absolute coefficient than general factors and particular narrators.
For example, the average coefficient of the top $k\%$ most positively correlated core contexts ranges from ($0.18$, $0.28$, $0.33$) to ($0.35$, $0.59$, $0.78$) when the dimension $C$ is ($2,3,4$).}
They are more confident in explaining the transferability.
For example, with the core context composed of $locatedIn(des,CA)$, $ListCar(car)$ and $BigCar(car)$,
whose coefficient is $0.35$, 
we can explain the positive transfer $\mathcal{F}_{(DL,ORD,LAX) \rightarrow (AA,ORD,SFO)}$ more confidently by \textit{``The carrier of both source and target learning domain belongs to big and list airline companies, and their destination airports are both located in California"}.

Figure~\ref{res:core} [Right] reports that ($19.9\%$, $11.6\%$, $4.8\%$) of all the ($2$, $3$, $4$)-dimension entailment subsets have significant correlation analysis with FTI (i.e., $\rho(\mathcal{X})<0.05$), 
while ($13.6\%$, $1.8\%$, $0.2\%$) are valid core contexts (i.e., $\rho(\mathcal{X}) <0.05$ and $\left\| \gamma(\mathcal{X}) \right\| \geq 0.1$).
\kr{On one hand, as the dimension increases, the percentage of valid core contexts significantly decreases.
On the other hand, the fact that a very large part of the entailment subsets have insignificant correlation analysis verifies that {\tt EearlyStop} in core context searching (Algorithm \ref{alg:ccs}) is effective.
For example, when the dimension of the current core context is $4$,
it avoids $95.2\%$ of the traversing for core contexts with higher dimension.
}

\vspace{-0.15cm}
\vspace{-0.2cm}

\noindent \textbf{Discussion and Lessons.}
The evaluation presents the explanatory \kr{evidence's} confidence and \kr{percentage (i.e., the probability   of being available as evidence).}
For confidence, we have core contexts $>$ general factors $d^{new}$ and $d^{obs}$ $>$ entailment narrators $>$ general factor $d^{inv}$,
while for percentage (of being available), we have general factors ($100\%$) $>$ entailment narrators ($30.4\%$ in average) $>$ core contexts ($15.6\%$ in average).
General factors can successfully explain any negative transfers, but fail to provide confident \kr{evidence} for positive transfers.
Core contexts, especially those with high dimensions, 
have very high confidence but the percentage decreases quickly as the dimension grows.
For both high confidence and availability, all
the three kinds of \kr{evidence} need to be used together.

\kr{
The evaluation also analyzes the positive impact of our techniques, 
including (i) root individual selection, which saves much computation but keeps high quality \kr{evidence}, 
(ii) entailment reasoning and external axiom importing, which enrich the \kr{evidence} and improve the percentage,
and (iii) early stop strategy in core context searching, which significantly reduces unnecessary searching.
}

\kr{The explanations lead to insights of feature transfer for users without ML expertise, and in turn allow them to further improve a transfer learning approach with more optimized settings.}
For a specific target domain, the explanations can answer the question of what to transfer by comparing the \kr{evidence} of different source learning domains.
Meanwhile, we can infer explanatory \kr{evidence for different features such as different Conv layers of a CNN architecture}.
Thus for a specific pair of source and target learning domains, 
we can answer the question of when to transfer by selecting a feature that maximizes the positive \kr{evidence}.

\vspace{-0.1cm}
\section{Related Work}
ML explanation has been studied for years, \kr{mainly including model interpretation (i.e., understanding how decisions are made) and prediction justification (i.e., justifying why a particular decision is good)} \cite{biran2017explanation}.
In this section, we first overview the above two aspects,
and then introduce the state-of-the-art in transfer learning explanation and human-centric explanation.
\vspace{0.03cm}

\noindent \textbf{Model Interpretation.}
Some ML models are inherently interpretable.
One type is sparse linear models such as Supersparse Linear Integer Models \cite{ustun2016supersparse}. 
\kr{These models' variable coefficients can present how much each variable contributes to the decision.}
Another type is rule-based models such as sparse Decision Tree \cite{wu2017beyond} and \kr{Bayesian Rule Lists \cite{letham2015interpretable}.
They can explain the decision inference procedure with internal probabilities and rules.}
 
To interpret black-box models,
visualization techniques have been applied.
For example, \cite{zeiler2014visualizing} visualized the hidden layer output of a CNN to understand the feature representation of data. 
\kr{For another example, 
\cite{jakulin2005nomograms} proposed the algorithm {\tt nomograms} to visualize Support Vector Machines.}
Recent advances in model interpretation include 
\kr{(i) attention mechanism for weighting the importance of different input parts \cite{qin2017dual},
(ii) reasoning-based consistent sample selection for stream learning \cite{chen2017learning},}
(iii) data distribution summarization with prototypes and criticisms \cite{kim2016examples}, 
etc.

\noindent \textbf{Prediction Justification.}
A specific prediction can be explained by evaluating the effect of each meaningful input variable \cite{biran2017human}.
It can be directly calculated in an interpretable model
or estimated with input isolation strategies 
\kr{
such as omitting a subset of input 
\cite{robnik2008explaining,martens2014explaining}. 
For complex and black-box models,
\cite{baehrens2010explain,ribeiro2016should} proposed to approximate them by multiple linear
models which have interpretable data representations and local fidelity.

Generating description text is another approach to justify predictions.
Recent advances include
(i) caption generation for visual decisions such as image classification \cite{hendricks2016generating},
(ii) text description of effective ML features \cite{biran2017human}, etc.
}

\noindent \textbf{Transfer Learning Explanation.}
Current studies on transfer learning explanation mainly lie in transferability analysis.
Problems like when and what to transfer
have been investigated in both theory and practice \cite{pan2010survey,weiss2016survey}.
Recent advances include 
(i) experimental quantification of the generality (transferable) and specificity (untransferable) of CNN feature  \cite{yosinski2014transferable},
(ii) theoretic justification of the relation between feature structure similarity and transferability \cite{liu2017understanding}, 
etc.

These attempts of transferability analysis definitely benefit ML experts, 
but will fail to explain the learned model or justify the prediction to common people.
The understanding to transferability is encoded in a machine understandable way (e.g., loss function) to enhance learning.
The explanations are neither represented in a human understandable format 
nor enriched with common sense knowledge. 
To the best of our knowledge, 
there are currently no studies for human-centric transfer learning explanation.
\vspace{0.03cm}

\noindent \textbf{Human-centric ML Explanation.}
Human-centric ML explanation aims at interpreting learned models or justifying predictions with background or common sense knowledge in a human understandable way \cite{biran2017human}.
Most of the current studies are based on corpuses.
\cite{hendricks2016generating} utilized external corpuses to generate captions to explain image classification decisions,
while \cite{biran2017human} used Wikipedia articles to describe effective features of a ML model.
Few studies utilize semantic data in human-centric explanation.
\cite{tiddi2014dedalo} proposed 
a framework to traverse Linked Data and use graph path commonalities to explain data clusters.

The current studies incorporate external knowledge,
but ignore expressive knowledge e.g., ontology and their reasoning capability.
It lacks a general knowledge representation and reasoning framework to utilize local ontologies and external knowledge bases for human-centric ML explanation.
This work bridges the above gap 
and is among the first to study human-centric transfer learning explanation.



\section{Conclusion and Outlook}
In this study, we address the problem of human-centric transfer learning explanation.
Our ontology-based framework exploits the reasoning capability and external knowledge bases like DBpedia to infer different kinds of human understandable explanatory \kr{evidence}, 
including general factors, particular narrators and core contexts.
\kr{It allows common users without ML expertise to have a good insight of positive transfers and negative transfers,
and further answer the questions of what to transfer and when to transfer for more optimized transfer learning settings.}
The quality of explanatory \kr{evidence, including the confidence and availability, and the effect of our methods,
are evaluated with US flight departure delay prediction, where features learned by CNNs are transferred.}
In the future work, we will exploit more efficient core context search algorithms and the impact of semantic expressivity,
with experiments in one more prediction application.





\section{Acknowledgments}
The work was partially funded by the project SIRIUS and the EPSRC project DBOnto.

\bibliographystyle{named}
\bibliography{XTL}
\end{document}